\pdfoutput=1
\documentclass[11pt]{article}

\usepackage{acl}

\usepackage{times}
\usepackage{latexsym}

\usepackage[T1]{fontenc}
\usepackage[utf8]{inputenc}

\usepackage{microtype}

\usepackage{url}
\usepackage{amssymb}
\usepackage{amsmath}
\usepackage{subfig}
\usepackage{graphicx}
\usepackage{graphics}
\usepackage{booktabs}
\usepackage{multirow}
\usepackage{color}
\usepackage{fixltx2e}
\usepackage{enumitem}
\usepackage{pgfplots}
\usepackage{makecell}
\usepackage{comment}
\usepackage{nicefrac}
\usepackage{amsfonts}
\usepackage{bm}
\usepackage{amsthm}

\usepackage{enumitem}
\setlist[itemize,enumerate]{leftmargin=*}
\usepackage[normalem]{ulem}
\usepackage{tikz}
\usepackage{xcolor}

\definecolor{fc}{HTML}{1E90FF}
\definecolor{h}{HTML}{228B22}
\definecolor{bias}{HTML}{87CEFA}
\definecolor{noise}{HTML}{8B008B}
\definecolor{conv}{HTML}{FFA500}
\definecolor{pool}{HTML}{B22222}
\definecolor{up}{HTML}{B22222}
\definecolor{view}{HTML}{FFFFFF}
\definecolor{bn}{HTML}{FFD700}
\tikzset{fc/.style={black,draw=black,fill=fc,rectangle,minimum height=1cm}}
\tikzset{h/.style={black,draw=black,fill=h,rectangle,minimum height=1cm}}
\tikzset{bias/.style={black,draw=black,fill=bias,rectangle,minimum height=1cm}}
\tikzset{noise/.style={black,draw=black,fill=noise,rectangle,minimum height=1cm}}
\tikzset{conv/.style={black,draw=black,fill=conv,rectangle,minimum height=1cm}}
\tikzset{pool/.style={black,draw=black,fill=pool,rectangle,minimum height=1cm}}
\tikzset{up/.style={black,draw=black,fill=up,rectangle,minimum height=1cm}}
\tikzset{view/.style={black,draw=black,fill=view,rectangle,minimum height=1cm}}
\tikzset{bn/.style={black,draw=black,fill=bn,rectangle,minimum height=1cm}}

\newcommand{\norm}[1]{\left\lVert #1 \right\rVert}

\title{\textsc{CometKiwi}: \\ IST-Unbabel 2022 Submission for the Quality Estimation Shared Task}

\usepackage[misc]{ifsym}

\author{
    Ricardo Rei\thanks{~~Equal contribution. \Letter \, \url{ricardo.rei@unbabel.com}}\,\,$^{1,2,4}$,
    Marcos Treviso$^{*3,4}$,
    Nuno M. Guerreiro$^{*3,4}$,
    Chrysoula Zerva$^{*3,4}$,
    \\
    \bf
    Ana C. Farinha$^{1}$,
    Christine Maroti$^{1}$,
    José G. C. de Souza$^{1}$,
    Taisiya Glushkova$^{3,4}$,\\
    \bf
    Duarte M. Alves$^{1,4}$,
    Alon Lavie$^{1}$,
    Luisa Coheur$^{2,4}$,
    André F. T. Martins$^{1,3,4}$
    \\
    $^{1}$Unbabel, Lisbon, Portugal, \,\ $^{2}$INESC-ID, Lisbon, Portugal \\
    $^{3}$Instituto de Telecomunicações, Lisbon, Portugal \\
    $^{4}$Instituto Superior Técnico, University of Lisbon, Portugal
}

\begin{document}
\maketitle
\begin{abstract}
We present the joint contribution of IST and Unbabel to the WMT 2022 Shared Task on Quality Estimation (QE). Our team participated
on all three subtasks: (i) Sentence and Word-level Quality Prediction; (ii) Explainable QE; and (iii) Critical Error Detection. 
For all tasks we build on top of the \textsc{Comet} framework, connecting it with the predictor-estimator architecture of \textsc{OpenKiwi}, and equipping it with a word-level sequence tagger and an explanation extractor.
% for sentence-level models. 
Our results suggest that incorporating references during pretraining improves performance across several language pairs on downstream tasks, and that jointly training with sentence and word-level objectives yields a further boost. 
Furthermore, combining attention and gradient information proved to be the top strategy for extracting good explanations of sentence-level QE models. 
Overall, our submissions achieved the best results for all three tasks for almost all language pairs by a considerable margin.%
\footnote{\url{https://github.com/Unbabel/COMET}}
\end{abstract}

\section{Introduction}

Quality Estimation (QE) is the task of automatically assigning a quality score to a machine translation output without depending on reference translations \cite{specia2018quality}. In this paper, we describe the joint contribution of Instituto Superior T\'ecnico (IST) and Unbabel to the WMT22 Quality Estimation shared task, where systems were submitted to three tasks: (i) Sentence and Word-level Quality Prediction; (ii) Explainable QE; and (iii) Critical Error Detection. 

This year, we leverage the similarity between the tasks of MT evaluation and QE and bring together the strengths of two frameworks, \textsc{Comet}~\citep{rei-etal-2020-comet}, which has been originally developed for reference-based MT evaluation, and \textsc{OpenKiwi}~\citep{kepler-etal-2019-openkiwi}, which has been developed for word-level and sentence-level QE. 
Namely, we implement some of the features of the latter, as well as other new features, into the \textsc{Comet} framework. The result is \textsc{CometKiwi}, which 
links the predictor-estimator architecture with \textsc{Comet} training-style,
and incorporates word-level sequence tagging.
% and explanation extraction. 
% as follows: ...
% With the successes of the \textsc{Comet} framework for MT evaluation in this year's submission we extended \textsc{Comet} 
% to support the predictor-estimator architecture, 
% alongside supporting word-level sequence tagging, and explanation extraction from sentence-level systems. 
%Also, following the traditional Predictor-Estimator architecture we redesign \textsc{Comet} encoders to support cross-encoding\footnote{Previous \textsc{Comet} QE models followed a dual-encoder architecture rather than encoding translations along with their respective sources (cross-encoding)}. 

% \andre{not all of it, right? some LPs had training data}
Given that some language pairs (LPs) in the test set were not present in the training data, we aimed at developing QE systems that achieve good multilingual generalization and that are flexible enough to account for unseen languages through few-shot training. 
% Since the test data was composed of LPs for which there is no training data (only a few examples as development data, with the exception of English-Yoruba, for which there is not even development data available), our 
To do so, we start by pretraining our QE models on Direct Assessments (DAs) annotations from the previous year's Metrics shared task as it was shown to be beneficial in our previous submission~\citep{zerva-etal-2021-ist}. Then we fine-tune our models with the data made available by the shared task.\footnote{For zero-shot LPs we use only the 500 training examples.} 
% Furthermore, we experimented with different types
% of uncertainty-related information to leverage it’s
% benefits, improving performance and robustness
% of the submitted systems (see §3.1.1). All related
% code extensions will be publicly available.
We experimented with different pretrained multilingual transformers as the backbones of our models, and we developed new explainability methods to interpret them. We describe our systems and their training strategies in Section~\ref{sec:implemented_systems}. 
% All related code extensions will be publicly available.
Overall, our main contributions are:
\begin{itemize}
    \item We combine the strengths of \textsc{Comet} and \textsc{OpenKiwi}, leading to \textsc{CometKiwi}, a model that adopts \textsc{Comet} training features useful for multilingual generalization along with the predictor-estimator architecture of \textsc{OpenKiwi}.
    % (e.g. discriminative learning rates, keeping parts of the model frozen during fine-tuning, layer combination through scalar mix, etc.).
    
    \item Following our previous work \citep{zerva-etal-2021-ist}, we show the importance of pretraining QE models on annotations from the Metrics shared task.
    
    \item We show that we can improve results for new LPs with only 500 examples without harming correlations for other LPs. 
    
    \item We propose a new interpretability method that uses attention and gradient information along with a head-level scalar mix module that further refines the relevance of attention heads.
    % of sentence-level QE models.
    % , leading to better explainability performance and allowing us to choose relevant attention heads at inference time for zero-shot LPs. 
\end{itemize}

\textbf{Our submitted systems achieve the best multilingual results
on all tracks by a considerable margin}: for sentence-level DA our system achieved a 0.572 Spearman correlation (+7\% than the second best system); for word-level our system achieved a 0.341 MCC score (+2.4\% than the second best system); and for Explainable QE our system achieved 0.486 R@K score (+10\% than the second best system). The official results for all LPs are presented in Table~\ref{tab:results_task_1_and_2} in the appendix.

% \andre{these sentences are a little bit convoluted and hard to understand. I would have a paragraph about the data (being more precise about which LPs have training data and which don't), followed by e.g. bullet points describing our procedure}. 

% \begin{itemize}
%     \item Introduction: In the intro we should add a table summarizing our results and highlighting that we won by significant margin on most tasks.
%     % \item Implemented Systems:
%     %     \subitem{Pretrained models} \ricardo{Ricardo}
%     %     \subitem{Task 1: Quality prediction -- includes Sentence and word-level DA/MQM} \ricardo{Ricardo, Chryssa, Catarina, Christine, Taya}
%     %     \subitem{Task 2: Explainable QE} \ricardo{Nuno, Marcos}
%     %     \subitem{Task 3: Critical Error Detection} \ricardo{Catarina}
%     \item Experimental Results
%         \subitem{Pretrained models} \ricardo{Ricardo}
%         \subitem{Task 1: Quality prediction -- includes Sentence and word-level DA/MQM} \ricardo{Ricardo, Chryssa, Catarina, Christine, Taya, Jose}
%         \subitem{Task 2: Explainable QE}  \ricardo{Nuno, Marcos}
%         \subitem{Task 3: Critical Error Detection} \ricardo{Catarina}
%     \item Conclusions 
% \end{itemize}

\section{Background}

% \andre{we don't need this level of detail here. there will be a findings paper describing QE; we only need to describe the details that are necessary to describe our system} \marcos{For the explainable QE it makes sense to define all this since we explicitly use attention heads to define our explainers.} \marcos{But if it takes too much space I guess we can trim or remove some parts?}

\paragraph{Quality Estimation.} QE systems are usually designed according to the granularity in which predictions are made, such as sentence and word-level. In sentence-level QE, the goal is to predict a single quality score $\hat{y} \in \mathbb{R}$ given the whole source and its translation as input. Word-level QE works in a lower granularity level, with the goal of predicting binary quality labels $\hat{y}_i \in \{\textsc{ok}, \textsc{bad}\}$ for all $1 \leq i \leq n$ \emph{machine-translated words}, indicating whether that word is a translation error or not.

\paragraph{Transformers.} 
The multi-head attention mechanism is the key component in transformers, being responsible for contextualizing the information within and across input sentences \citep{vaswani2017attention}. Concretely, given as input a matrix $\bm{Q} \in \mathbb{R}^{n \times d}$ containing $d$-dimensional representations for $n$ queries, and matrices $\bm{K},\bm{V} \in \mathbb{R}^{m \times d}$ for $m$ keys and values, the \textit{scaled dot-product attention} at a single head is computed as:
\begin{equation}\label{eq:dotproduct-attention}
    \textsf{att}(\bm{Q}, \bm{K}, \bm{V}) = 
    % \mathrm{softmax}
    \pi
    \underbrace{\Bigg(
        \frac{\bm{Q}\bm{K}^\top}{\sqrt{d}}
    \Bigg)}_{\bm{Z} \in \mathbb{R}^{n \times m}} 
    \bm{V} \in \mathbb{R}^{n \times d}.
\end{equation}
The $\pi$ transformation maps rows to distributions, with \textsf{softmax} being the most common choice, 
$\pi(\bm{Z})_{ij} = \textsf{softmax}(\bm{z}_i)_j$. 
% In words, the attention mechanism computes for each query a weighted representation of the values.
Multi-head attention is computed by evoking Eq.~\ref{eq:dotproduct-attention} in parallel for each head $h$: 
\begin{equation}\label{eq:multihead-attention} \nonumber
    \textsf{head}_h(\bm{Q}, \bm{K}, \bm{V}) = \textsf{att}(\bm{Q}\bm{W}^Q_h, \bm{K}\bm{W}^K_h, \bm{V}\bm{W}^V_h),
\end{equation}
where $\bm{W}^Q_h$, $\bm{W}^K_h$, $\bm{W}^V_h$ are learnable linear transformations. 
Finally, the output of the multi-head attention module at the $\ell$-th layer is a set of hidden states $\bm{H}_{\ell} \in \mathbb{R}^{n \times d}$ formed via the concatenation of all $\bm{h}_{\ell,1}, ..., \bm{h}_{\ell,H}$ heads in that layer followed by a learnable linear transformation $\bm{W}^O$:
\begin{equation}\label{eq:multihead-attention-output} \nonumber
    \bm{H}_{\ell} = \textsf{concat}(\bm{h}_{\ell,1}, ..., \bm{h}_{\ell,H})\bm{W}^O.
\end{equation}
% This way, heads have the capability of learning specialized phenomena. 
The hidden states are further refined through position-wise feed-forward blocks and residual connections to obtain a final representation: $\bm{H}_{\ell} = \mathsf{FFN}(\bm{H}_{\ell}) + \bm{H}_{\ell}$.
% \nuno{Are we ignoring layer normalization here?} \marcos{yes}
Transformers with only encoder-blocks, such as BERT \citep{devlin-etal-2019-bert} and XLM \citep{conneau-etal-2020-unsupervised}, have only the encoder self-attention, and thus $m=n$.

\section{Implemented Systems} \label{sec:implemented_systems}

% \looseness=-1

\definecolor{paired-light-blue}{RGB}{198, 219, 239}
\definecolor{paired-dark-blue}{RGB}{49, 130, 188}
\definecolor{paired-light-orange}{RGB}{251, 208, 162}
\definecolor{paired-dark-orange}{RGB}{230, 85, 12}
\definecolor{paired-light-green}{RGB}{199, 233, 193}
\definecolor{paired-dark-green}{RGB}{49, 163, 83}
\definecolor{paired-light-purple}{RGB}{218, 218, 235}
\definecolor{paired-dark-purple}{RGB}{117, 107, 176}
\definecolor{paired-light-gray}{RGB}{217, 217, 217}
\definecolor{paired-dark-gray}{RGB}{99, 99, 99}
\definecolor{paired-light-pink}{RGB}{222, 158, 214}
\definecolor{paired-dark-pink}{RGB}{123, 65, 115}
\definecolor{paired-light-red}{RGB}{231, 150, 156}
\definecolor{paired-dark-red}{RGB}{131, 60, 56}
\definecolor{paired-light-yellow}{RGB}{231, 204, 149}
\definecolor{paired-dark-yellow}{RGB}{141, 109, 49}
\tikzset{%
    layernode/.style = {
        align=center,
        text width=6cm,
        inner sep=0.25cm,
        outer sep=0cm,
        rounded corners=4pt,
        fill=paired-light-gray!30,
        draw=paired-dark-gray!50,
    },
    halfnode/.style = {
        align=center,
        text width=2.5cm,
        inner sep=0.25cm,
        outer sep=0cm,
        rounded corners=4pt,
        fill=paired-light-gray!30,
        draw=paired-dark-gray!50,
    },
    outputnode/.style = {
        align=center,
        text width=2.6cm,
        % inner sep=0.25cm,
        % outer sep=0cm,
        % rounded corners=4pt,
        % fill=paired-light-orange!30,
        % draw=paired-dark-orange!45,
    },
    inputnode/.style = {
        align=center,
        % rounded corners=3pt,
        inner sep=0.2cm,
    },
}

\begin{figure}[t]
    \centering
    \small
    \begin{tikzpicture}
        %Nodes
        \node[inputnode] (input) at (0, 0) {
            \texttt{
            [cls] \textcolor{paired-dark-blue}{\textbf{target}}  [sep] \textcolor{paired-dark-red}{\textbf{source}} [eos]
            }
        };
        \node[layernode] (encoder) at (0, 1) {Pre-trained Encoder};
        \node[layernode] (scalarmix) at (0, 2) {Scalar Mix};
        % \node[halfnode] (avgpool) at (-1.7, 3) {Avg. Pooling};
        \node[halfnode] (avgpool) at (-1.7, 3) {\texttt{[cls]}};
        \node[halfnode] (piece) at (1.7, 3) {First Piece Select.};
        \node[halfnode] (ffn1) at (-1.7, 4) {Feed Forward};
        \node[halfnode] (ffn2) at (1.7, 4) {Feed Forward};
        \node[outputnode] (out1) at (-1.7, 5) {Sentence score \\ $\hat{y} \in \mathbb{R}$};
        \node[outputnode] (out2) at (1.7, 5) {Word labels \\ $\hat{y}_i \in \{\textsc{ok}, \textsc{bad}\}$};
        
        %Lines
        \draw[->] (input) to (encoder);
        \draw[->] (encoder) to (scalarmix);
        % \draw[->] (scalarmix) to (avgpool);
        % \draw[->,paired-dark-red] (-0.5, 2.35) to (-0.5, 2.6);
        % \draw[->,paired-dark-red] (-1.0, 2.35) to (-1.0, 2.6);
        % \draw[->,paired-dark-red] (-1.5, 2.35) to (-1.5, 2.6);
        \draw[->] (-1.7, 2.35) to (-1.7, 2.6);
        % \draw[->,paired-dark-blue] (-2.0, 2.35) to (-2.0, 2.6);
        % \draw[->,paired-dark-blue] (-2.5, 2.35) to (-2.5, 2.6);
        % \draw[->,paired-dark-blue] (-3.0, 2.35) to (-3.0, 2.6);
        % \draw[->] (scalarmix) to (piece);
        \draw[->,paired-dark-blue] (0.5, 2.35) to (0.5, 2.6);
        \draw[->,paired-dark-blue] (1.0, 2.35) to (1.0, 2.6);
        \draw[->,paired-dark-blue] (1.5, 2.35) to (1.5, 2.6);
        \draw[->,paired-dark-blue] (2.0, 2.35) to (2.0, 2.6);
        \draw[->,paired-dark-blue] (2.5, 2.35) to (2.5, 2.6);
        \draw[->,paired-dark-blue] (3.0, 2.35) to (3.0, 2.6);
        \draw[->] (avgpool) to (ffn1);
        \draw[->] (piece) to (ffn2);
        \draw[->] (ffn1) to (out1);
        \draw[->] (ffn2) to (out2);

    \end{tikzpicture}
    
    \caption{General architecture of \textsc{CometKiwi} for sentence-level (left part) and word-level QE (right part).}
    \label{fig:arch}
\end{figure}
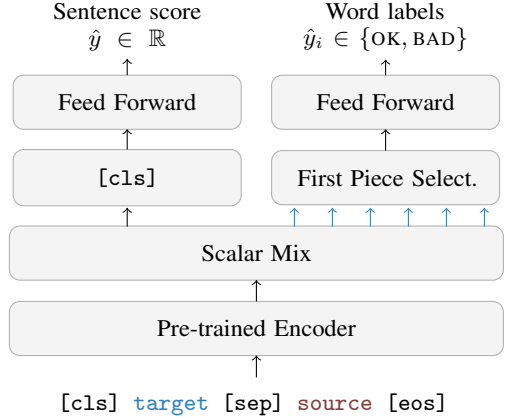

The overall architecture of our models is shown in Figure~\ref{fig:arch}. 
The machine translated sentence $\bm{t} = \langle t_1, ..., t_n \rangle$ and its source sentence counterpart $\bm{s} = \langle s_1, ..., s_m \rangle$
% , and the reference translation $\bm{r} = \langle r_1, ..., r_q \rangle$ 
are concatenated and passed as input to the encoder, which produces $d$-dimensional hidden state vectors
% \andre{actually each $\bm{H}_i$ is a matrix whose columns are $d$-dimensional hidden state vectors...} \marcos{Yup. I think the notation is ok here}
$\bm{H}_0, ..., \bm{H}_{L}$ for each layer $0 \leq \ell \leq L$, where $\bm{H}_i \in \mathbb{R}^{(n+m)\times d}$, where $\ell=0$ corresponds to the embedding layer. Next, all hidden states are fed to a scalar mix module~\citep{peters-etal-2018-deep} that learns a weighted sum of the hidden states of each layer of the encoder, producing a new sequence of aggregated hidden states $\bm{H}_{\mathrm{mix}}$ as follows:
\begin{equation}\label{eq:scalarmix}
    \bm{H}_{\mathrm{mix}} = \lambda \sum_{\ell = 0}^{L} \beta_{\ell} \bm{H}_{\ell},
\end{equation}
where $\lambda$ is a scalar trainable parameter, $\bm{\beta} \in \triangle^L$,  
% \andre{define $\triangle^L$; btw I prefer the notation $\triangle_L$, but in this case if we're not using it elsewhere you can just say $\bm{\beta}$ is a probability vector satisfying $\bm{\beta} \ge \mathbf{0}$ and $\mathbf{1}^\top \bm{\beta} = 1$} \marcos{I think $$\triangle^{L-1}$ makes more sense as it is the constrained space within $\mathbb{R}^L$.}
is given by $\bm{\beta} = \textsf{sparsemax}(\bm{\phi})$ using a sparse transformation~\citep{martins2016softmax}, with $\bm{\phi} \in \mathbb{R}^L$ as learnable parameters and $\triangle^L := \{\bm{\beta} \in \mathbb{R}^L: \mathbf{1}^\top \bm{\beta} = 1, \bm{\beta} \ge 0\}$\footnote{As it has been shown in \cite{rei-etal-2022-searching} not all layers are relevant and thus, using \textsf{sparsemax} we learn to ignore layers that do not help in the task at hands}. 
% \marcos{the default scalarmix uses softmax, and it is the one that we used for Task 1, right?}\ricardo{I think I ended up using sparsemax for most models.} \marcos{ok}
% \looseness=-1

For sentence-level models, the hidden state of the first token (\texttt{<cls>}) is used as sentence representation $\bm{H}_{\mathrm{mix}, 0} \in \mathbb{R}^{d}$, which, in turn, is passed to a 2-layered feed-forward module in order to get a sentence score prediction $\hat{y} \in \mathbb{R}$. For word-level models, we first retrieve the hidden state vectors associated with the first word piece of each machine translated token, and then pass them to a linear projection to get word-level predictions $\hat{y}_i \in \{\textsc{ok}, \textsc{bad}\},\, \forall_{1 \leq i \leq n}$.
Moreover, attention matrices $\bm{A}_{1,1}, ..., \bm{A}_{L,H}$ for all layers and heads are also recovered as a by-product of the forward propagation. \looseness=-1

\paragraph{Pretraining on Metrics Data.}
% \label{sec:pretraining}
Every year, the WMT News Translation shared task organizers collect human judgments in the form of DAs. The collective corpora of 2017, 2018, and 2019 contain 24 LPs and a total of 657k samples with source, target, reference, and DA score. We follow our experiments from last year \cite{zerva-etal-2021-ist} and start by pretraining our QE models on this data using the learning objective proposed by UniTE~\cite{wan-etal-2022-unite}, which incorporates reference translations into training and thus acts as data augmentation.

%given that, for each samples, we learn to predict $y_i$ by looking at source, reference and source + reference.
%\begin{equation}
%    \mathcal{L} = \mathcal{L}_{ref} + \mathcal{L}_{src} + \mathcal{L}_{src+ref}
%\end{equation}

%given by concatenate the machine translated, source, and reference sentences using the template \texttt{<s> mt </s> <s> src </s> <s> ref </s>} for InfoXLM, and \texttt{<cls> mt <sep> src <sep> ref <eos>} for RemBERT.

\paragraph{Setting pretrained transformers as encoders.} 
We follow the recent trend \citep{kepler-etal-2019-openkiwi,ranasinghe-etal-2020-transquest-wmt2020} and experiment with three different pretrained multilingual transformers as the encoder layer of our models: 
XLM-R Large~\citep{conneau-etal-2020-unsupervised},\footnote{\url{https://huggingface.co/xlm-roberta-large}} InfoXLM Large~\citep{chi-etal-2021-infoxlm},\footnote{\url{https://huggingface.co/microsoft/infoxlm-large}} 
and RemBERT~\citep{chung2021rethinking}.\footnote{\url{https://huggingface.co/google/rembert}} 
XLM-R and InfoXLM consist of 24 encoder blocks with 16 attention heads each, whereas RemBERT has 32 encoder blocks with 18 attention heads each. 
% \nuno{This paragraph is heavily duplicated from the first paragraph of this section. We can remove the types of encoders from there, and only mention them here.} \marcos{done!}

% \paragraph{XLM-R as encoder.} We set a XLM-R Large \citep{conneau-etal-2020-unsupervised} as the encoder layer.\footnote{\url{https://huggingface.co/xlm-roberta-large}}  
% XLM-R is a pretrained multilingual transformer trained with the Masked language modeling (MLM) objective introduced in BERT \cite{devlin-etal-2019-bert}.

% \paragraph{InfoXLM as encoder.} We set a InfoXLM Large \citep{chi-etal-2021-infoxlm} as the encoder layer.\footnote{\url{https://huggingface.co/microsoft/infoxlm-large}}  
% InfoXLM is a pretrained cross-lingual transformer trained to maximize mutual information between multilingual-multi-granularity texts. It consists of 24 encoder blocks with 16 attention heads each. 

% \paragraph{RemBERT as encoder.} We replace the InfoXLM by a RemBERT model as the encoder layer \citep{chung2021rethinking}.\footnote{\url{https://huggingface.co/google/rembert}} 
% RemBERT is a pretrained multilingual transformer that can be seen as a larger multilingual BERT with decoupled input and output embeddings, which helps to accelerate training. It consists of 32 encoder blocks with 18 attention heads each.

\subsection{Task 1: Quality prediction}

After the pretraining phase,
% described in \S\ref{sec:pretraining}, 
we adapt our models to the released QE data using source and translation (i.e., in this phase we do not include references) to the different type of quality assessments provided, namely, DA and HTER\footnote{HTERs are available only for word-level subtasks.} from the MLQE-PE corpus~\cite{fomicheva2020mlqepe} and MQM annotations from WMT 2020 and 2021~\cite{freitag-etal-2021-experts, freitag-etal-2021-results}.

\subsubsection{Sentence-level quality prediction}

For the sentence-level QE task we consider a multi-task setting (using sentence scores alongside supervision from \textsc{ok}/\textsc{bad} tags) and the sentence-level only setting, with supervision only from the sentence-level quality assessment $y$. We found that adding the word-level supervision was beneficial for models built on top of InfoXLM. For the sentence-level supervision we used both DA and MQM scores. In this multi-task setting we use a combined loss as described in Eq. \ref{eq:comb_loss}: 
\begin{align}
    % old:
    % \mathcal{L} = \lambda \frac{1}{N}\sum_{i=0}^{N}(y_{\mathrm{s}},\hat{y}_\mathrm{s})^2 + (1-\lambda)\sum_{y_{\mathrm{w}} \in \mathcal{W}} w(y_{\mathrm{w}}) \log \hat{p}(y_{\mathrm{w}})
    % divided:
    \mathcal{L}_{\mathrm{sent}}(\theta) &= \frac{1}{2}(y - \hat{y}(\theta))^2 \\
    \mathcal{L}_{\mathrm{word}}(\theta) &= -\frac{1}{n}\sum_{i=1}^{n} w_{y_i} \log p_\theta(y_i) \\
    \mathcal{L}(\theta) &= \lambda_s \mathcal{L}_{\mathrm{sent}}(\theta) + \lambda_w \mathcal{L}_{\mathrm{word}}(\theta), \label{eq:comb_loss}
    % all in one:
    % \mathcal{L} &= \lambda_s (y - \hat{y})^2 + \frac{\lambda_w}{n}\sum_{i=1}^{n} -w_{y_i} \log \hat{p}(y_i), \label{eq:comb_loss}
\end{align}
where $w \in \mathbb{R}^2$ represents the class weights given for \textsc{ok} and \textsc{bad} tags, and $\lambda_s, \lambda_w$ are used to weigh the combination of the sentence and word-level losses, respectively. Note that $\lambda_s = 1$ and $\lambda_w = 0$ yields a fully sentence-level model.

\paragraph{Few-shot language adaptation.} Since in this shared task submissions are tested on 5 LPs for which there is no official training data (\textit{km-en}, \textit{ps-en}, \textit{en-ja}, \textit{en-cs}, \textit{en-yo}), we experimented with few-shot adaptation using half of the data released in the official development set. The official development set has 1K examples for each language pair (except \textit{en-yo} for which there is no available data). To perform few-shot language adaptation we split the data into two halves: one for fine-tuning and another for validation.

\paragraph{Ensembling models.} For our final submission for Direct Assessments we combine six multilingual systems using different hyperparameters by computing an weighted average of their outputs, where the weights for each language pair were tuned with Optuna~\citep{optuna_2019}. The major difference between the ensembled models comes from the underlying encoder and whether or not they used word-level supervision. Three models of our final ensemble use word-level supervision while the other three use only sentence-level supervision. Regarding the encoder, three models use InfoXLM, two models use RemBERT and a single model uses XLM-R.

Our final submission for MQM predictions was an ensemble of eleven multilingual systems, which combined the six systems used in the DA ensemble as well as five additional systems.
% \footnote{One of the 5 models -- a system fine-tuned only on DAs for the 3 MQM LPs  -- had only 0.02 weight for English-Russian and 0 weight for the other two LPs, suggesting it was redundant to include in the ensemble.}
For these additional systems, we made two major adjustments to the fine-tuning process. First, we filtered the DA data to the languages that were included in the MQM LPs, namely \textit{ru-en}, \textit{en-zh}, and \textit{en-de}. Second, we incorporated the MQM data into the fine-tuning process, either as an additional fine-tuning step after fine-tuning on the language-filtered DA data, or by concatenating the DA and MQM data together. All additional systems used word-level supervision in addition to sentence-level and used InfoXLM as encoder.

\subsubsection{Word-level quality prediction}
Similarly, for the word-level QE tasks we experimented with both the multi-task setting and word-labels only ($\lambda_s = 0$ and $\lambda_w = 1$). Overall, we found that adding the sentence-level supervision was beneficial, especially for the languages pairs included in the test-set. Nonetheless, for some LPs, ignoring sentence-level supervision showed superior performance. 
%For the sentence level supervision we used both DA, HTER and MQM scores. 
Due to the mix of high-, mid- and low-resource languages in the data, the distribution of \textsc{ok} and \textsc{bad} tags differs substantially between LPs leading to inconsistent performance in terms of MCC (see Table~\ref{tab:qe_data} in the appendix). To mitigate this, for the word-level subtask, we prepend a language prefix token to the beginning of the source and target segments during training and testing. 

\paragraph{Pretraining on post-edit corpora.} Extending the pretraining on Metrics data, we pretrain the word-level models on two corpora that include both word-level labels and sentence (HTER) scores, namely QT21 \cite{specia2017translation} and APEQuest \cite{ive-etal-2020-post}. We compute the sentence-level score, using translation edit rate (TER) \cite{Snover06astudy} between the target and the corresponding post-edited sentence.

\paragraph{Ensembling models.} For word-level we followed a similar ensembling technique used for sentence-level, namely we combine multiple systems trained with different hyperparameters, encoders and pre-training setups. In the case of word-level predictions however, we need to resolve how to aggregate multiple predictions into OK/BAD tags. We use Optuna~\citep{optuna_2019} to choose how to weight and combine the models based on performance for each language pair on our internal test-set and we compare three different approaches:
\begin{enumerate}
    \item A naive ``best-only'' approach: we identify the best model for each LP and use its predictions.
    \item We ensemble the logits of each model: for each input segment we compute an ensembles of logits as $\sum_{i \in \mathcal{M}} w_i v_i$, where $\mathcal{M}$ is the set of models, $w_i$ is the weight of each model and $v_i$ the model logit vector. We use Optuna to find the optimal weight $w_i$ for each model in each LP.
    \item We ensemble the predicted tags of each model: for each input segment we compute an ensembles of tags as $\alpha \sum_{i \in \mathcal{M}} w_i c_i$, where $c_i$ is the predicted class and $\alpha$ is the weight given for the \textsc{bad} class. We use Optuna to find the optimal weights $w_i$ for each model and the optimal \textsc{bad} weight $\alpha$ for each LP.
\end{enumerate}
  
In the final submission we combine five models for the post-edit originated LPs: a RemBERT based model, an InfoXLM based model pretrained on APEQuest and QT21, and three checkpoints that are based on InfoXLM but use different parameters for the \textsc{bad/ok} weights and learning rate that were found via Optuna. 
For MQM we also combine five models, but this time instead of choosing three checkpoints based on optimising weights and learning rate, we use three different checkpoints with different training data mix on the relevant DA LPs, as this seemed to impact the performance on MQM word-level more than the weight ratios. Refer to \S\ref{sec:experiments} and Table \ref{tab:results_task_1_word_level_da} for more details.
% \ricardo{Chryssa maybe you should add a bit more detail here, similarly to what i did for sentence-level} \marcos{If it was the same techinique, I think this is not necessary.}

\begin{table*}[t]
    \centering
    \small
    \setlength{\tabcolsep}{.4em}
    \begin{tabular}{lccccccccccccc}
        \toprule
        & \multicolumn{12}{c}{\bf Direct Assessment} \\
        \cmidrule{2-14}
        \bf Encoder & \bf km-en & \bf ps-en & \bf en-ja & \bf en-cs &	\bf en-mr & \bf ru-en &	\bf ro-en &	\bf en-zh &	\bf en-de &	\bf et-en &	\bf si-en &	\bf ne-en &	\bf avg. \\
        \midrule
        \multicolumn{14}{c}{\textit{Baseline \cite{zerva-etal-2021-ist}}} \\
         XLM-R & 0.615 & 0.601 & 0.295 & 0.535 & 0.419 & 0.703 & 0.828 & 0.513 & 0.500 & 0.806 & 0.565 & 0.793 & 0.598 \\\midrule
        \multicolumn{14}{c}{\textit{Pretrained models}} \\
        InfoXLM  & 0.619	& 0.603 & 0.328 & 0.510 & 0.462 & 0.731	& 0.829	& 0.554 & 0.516 &	0.803 &	0.561 & 0.777 & 0.608 \\
        RemBERT  & 0.600 & 0.621	& 0.338	& 0.525	& 0.447 & 0.680 & 0.818 & 0.487 & 0.491 & 0.810 & 0.525 & 0.747 & 0.591 \\
        XLM-R  & 0.610 &	0.579 &	0.325 &	0.503 &	0.405 & 0.715 & 0.832 & 0.541 & 0.514 & 0.782 & 0.540 & 0.740 & 0.591 \\\midrule
        \multicolumn{14}{c}{\textit{Sentence-level only}} \\
        XLM-R & 0.628 & 0.591 & 0.350 & 0.531 & 0.551 & 0.761	& 0.859 & 0.577 & 0.568 & 0.800 & 0.565 &	0.796  & 0.631 \\
        InfoXLM & 0.629 & 0.623 & 0.348 & 0.515	& 0.574 & 0.747 &	0.858 & 0.586 & 0.551 &	0.828 &	0.568 &	0.790 & 0.635 \\
        RemBERT & 0.634 & 0.631 & 0.346 & 0.570 & 0.564 & 0.754 & 0.862 & 0.534 & 0.531 & 0.822 & 0.550 & 0.782 & 0.632 \\
        \multicolumn{14}{c}{\textit{Few-shot Language Adaptation}} \\
        XLM-R & 0.650 & 0.619 &	0.352 &	0.551 &	0.546 &	0.753 &	0.852 &	0.571 &	0.554 &	0.813 &	0.562 &	0.798 & 0.635 \\
        InfoXLM & 0.641 & 0.650 & 0.367 & 0.549 & 0.549 & 0.751 & 0.855 & 0.591 & 0.565 & 0.824 & 0.563 & 0.803 & 0.642 \\
        RemBERT & 0.625 & 0.641 & 0.367 & 0.568 & 0.563 & 0.756 & 0.857 & 0.540 & 0.527 & 0.824 & 0.568 & 0.796 & 0.636 
        \\\midrule
        \multicolumn{14}{c}{\textit{Sentence + word-level training}} \\
        InfoXLM & 0.617 & 0.586 & 0.344 & 0.532 & 0.572 & 0.761 & 0.865 & 0.586 & 0.579 & 0.829 & 0.576 & 0.804 & 0.637 \\
        RemBERT & 0.634 &	0.628 &	0.356 &	0.564 &	0.571 &	0.762 &	0.860 &	0.541 &	0.553 &	0.826 &	0.564 &	0.799 &	0.638 \\
        \multicolumn{14}{c}{\textit{Few-shot Language Adaptation}} \\
        InfoXLM & 0.643 & 0.632 & 0.335 & 0.557 & 0.560 & 0.766 & 0.860 & 0.575 & 0.582 & 0.833 &	0.578 &	0.809 & 0.644 \\
        RemBERT & 0.644	& 0.645 & 0.356 & 0.567 & 0.568 & 0.759 & 0.856 & 0.545 & 0.552 & 0.835 & 0.561 & 0.804 & 0.641 \\\midrule
        \multicolumn{14}{c}{\textit{Final Ensemble}} \\
        Ensemble 6x & \bf 0.664 & \bf 0.669 & \bf 0.380 & \bf 0.591 & \bf 0.593 & \bf 0.782 & \bf 0.871 & \bf 0.597 & \bf 0.593 & \bf 0.845 & \bf 0.588 & \bf 0.820 & \bf 0.666\\
        \bottomrule
    \end{tabular}
    \caption{Results for sentence-level QE in terms of Spearman correlation for DA.}
    \label{tab:results_task_1_DA}
\end{table*}

\subsection{Task 2: Explainable QE}

The goal of the Explainable QE task is to identify machine translation errors without relying on word-level label information. In other words, it can be cast as an unsupervised word-level quality estimation problem, where explanations can be seen as highlights, representing the relevance of input words w.r.t. the model's prediction via continuous scores, aiming at identifying tokens that were not properly translated. 

Several explainability methods can be used to extract highlights from a sentence-level model, such as post-hoc \citep{ribeiro2016should,arras-etal-2016-explaining} or inherently interpretable methods \citep{lei-etal-2016-rationalizing,guerreiro-martins-2021-spectra}. In our submission, we opted to use attention-based methods as they achieved the best results in the previous constrained track of the Explainable QE shared task  \citep{fomicheva-etal-2021-eval4nlp}. Concretely, we take inspiration in the method developed by \citet{treviso-etal-2021-ist}, which consists of scaling attention weights by the $\ell_2$-norm of value vectors \citep{kobayashi-etal-2020-attention} and finding the attention heads with the best performance on the dev set, and propose two new modifications:

\begin{itemize}
    \item \textbf{Attention $\times$ GradNorm:} Following the findings of \citet{chrysostomou-aletras-2022-empirical}, we decided to extract explanations that consider both attention and gradient information. More precisely, we scale the attention weights by the $\ell_2$-norm of the gradient of value vectors: 
    \begin{equation}
        \bm{A}_{\ell,h} \norm{ \nabla_{\bm{V}_{\ell, h}} }_2.
    \end{equation}
    
    \item \textbf{Head Mix:} We reformulate the scalar mix module (Eq.~\ref{eq:scalarmix}) to consider different weights for representations coming from different attention heads as follows:
    \begin{equation}
        \bm{H}_{\mathrm{mix}} = \lambda \sum_{\ell = 0}^{L} \beta_\ell \sum_{h=1}^{H} \gamma_{\ell, h} \bm{h}_{\ell, h},
    \end{equation}
    where the \textit{layer} mix coefficients $\bm{\beta} \in \triangle^L$ are given by $\bm{\beta} = \pi(\bm{\phi})$, and the \textit{head} mix coefficients $\bm{\gamma}_\ell \in \triangle^H$ are given by $\bm{\gamma}_\ell = \pi(\bm{\theta}_\ell)$. $\lambda \in \mathbb{R}$, $\bm{\phi} \in \mathbb{R}^L$ and $\bm{\theta} \in \mathbb{R}^{L \times H}$ are learnable parameters. We experimented both with dense ($\pi$ as \textsf{softmax}) and sparse ($\pi$ as \textsf{sparsemax}, \citealt{martins2016softmax}) transformations. After training, the Head Mix coefficients can help to find attention heads with high validation performance, which is helpful for explaining zero-shot LPs.
\end{itemize}

Furthermore, since all of our sentence-level models use subword tokenization, to get explanations for an entire word we follow \citet{treviso-etal-2021-ist} and sum the scores of its word pieces.

\paragraph{Ensembling explanations.} In our final submissions we average the explanation scores of different attention heads and layers to create a final explainer. We decided which heads and layers to aggregate together by looking at their performance on the dev set, selecting the top-5 with the highest explainability score.

\subsection{Task 3: Critical Error Detection}

Critical translations are defined as translations with strongly semantic deviations from the original source sentence, with the potential to lead to negative impacts in critical applications. The goal of this task is to predict sentence-level scores indicating whether a translation contains a critical error. Since the evaluation metrics automatically account for different binarization thresholds to separate good translations from bad ones, for this task we employed a single sentence-level InfoXLM model from Task 1 that was trained on DA data. Moreover, we participated only in the \emph{constrained setting}, meaning that we did not trained our systems specifically for this task. Therefore, our goal for this task was to validate whether our QE system from Task 1 was able to detect and differentiate translations with critical errors.

\section{Experimental Results}
\label{sec:experiments}
As we have seen in Section~\ref{sec:implemented_systems}, for our experiments we split the provided development sets into two equal size halves creating a new internal devset and an internal testset. The resulting sets contain $\approx$ 500 segments per language pair for both DA and MQM, word and sentence-level.
As for baselines we used our submitted systems from previous shared tasks: for Task 1 we used the \textsc{M1M-adapt}~\cite{zerva-etal-2021-ist}, and for Task 2 we used the $\text{Attn} \times \text{Norm}$ explainer~\cite{treviso-etal-2021-ist}.  The official results for Task 1 and Task 2 are shown in Table~\ref{tab:results_task_1_and_2}.
% \ricardo{Marcos can you double check?} \marcos{Yep. Looks neat}

\subsection{Quality Estimation}

Sentence-level submissions were evaluated using the Spearman’s rank correlation. Pearson’s correlation, MAE, and RMSE were also used as secondary metrics, but here we report only Spearman correlation since it was the primary metric used to rank systems. Word-level submission were evaluated using MCC, $F_1$-OK, and $F_1$-BAD, but we report only MCC as it was considered the main metric.
The submitted systems were independently evaluated on in-domain and zero-shot LPs for direct assessments and MQM.  

\paragraph{Direct Assessments.} Results for sentence-level DAs can be seen in Table~\ref{tab:results_task_1_DA}. The results show that the training strategies employed in \textsc{CometKiwi}, namely (i) pretraining models using Metrics data and (ii) incorporating references into training, lead to a correlation close to our best system from last year while disregarding the data from the MLQE-PE corpus. 
When fine-tuning on MLQE-PE data, we get overall improvements of $\sim4\%$, and further fine-tuning on new LPs gives $\sim1\%$ overall improvement. Still, for the unseen LPs (\textit{km-en, ps-en, en-ja, en-cs}), we got improvements between 2-3\% with just 500 samples. 
Among the three backbone transformers, we noticed that InfoXLM is the one that leads to a higher Spearman correlation (+1.7\% than XLM-R and RemBERT).
Furthermore, including word-level supervision always maintains or improves the results, especially for InfoXLM. In contrast, RemBERT does not seem to benefit from this signal.
% We observed that, for the sentence-level DA subtask, word-level supervision leads to better results for InfoXLM but does not seem to benefit RemBERT. Nonetheless, including word-level supervision always maintains or improves the results. 
We suspect that, for this task, the benefit of word-level supervision is not higher because the word-level information is coming from post-editions, which are conceptually different from DA annotations.

%     \item Fine-tuning of MLQE-PE leads to overall improvements of $\approx$ 4\% and further fine-tuning on the new LPs gives less than a 1\% overall improvement. Nonetheless, for the unseen LPs, we can get improvements between 2-3\% with just 500 samples.
%     \item We observed that, for the sentence-level DA subtask, word-level supervision leads to better results for InfoXLM but does not seem to benefit the RemBERT model. Nonetheless, word-level supervision always leads to equally good or better results than sentence-only supervision. We suspect that, for this task, the benefit is not larger because the word-level information is coming from PE labels which are different from DA annotations.
% \end{itemize}

% \marcos{Todo: break the itemize into a paragraph}

\paragraph{MQM.} Results for sentence-level MQM systems are shown in Table~\ref{tab:results_task_1_MQM}. The results show that the two main techniques used for adapting to MQM data, filtering DA data to the three MQM LPs and using MQM data for fine-tuning, improved Spearman correlations for all LPs over the pure DA baseline, for both sentence-level and multi-task systems. However, these techniques improved certain LPs more than others, so combining them together improved multilingual scores even further. 
Overall, we noticed that our results for MQM data have a high variance. To mitigate this, we concatenated the DA and MQM datasets together for a single fine-tuning, resulting in our best individual system on our internal test set. 
Due to these peculiarities in the MQM LPs, we decided to ensemble systems tuned on both DA and MQM data. Our final ensemble did not have as strong results as the individual systems on our internal test set, yet, it showed superior performance upon submission to codalab leader-board.

\begin{table}[t]
    \centering
    \small
    \setlength{\tabcolsep}{.4em}
    \begin{tabular}{lcccc}
        \toprule
        & \multicolumn{4}{c}{\bf MQM} \\
        \cmidrule{2-5}
        \bf System (fine-tuned on) & \bf en-de & \bf en-ru & \bf zh-en & \bf avg. \\
        \midrule
        \multicolumn{5}{c}{\textit{Sentence-level only}} \\
        DA & 0.529 & 0.534 & 0.215 & 0.426 \\
        DA + MQM & 0.531 & 0.552 & 0.250 & 0.444 \\
        DA (3 LPs) + MQM & 0.538 & 0.550 & 0.262 & 0.450 \\\midrule
        \multicolumn{5}{c}{\textit{Sentence + word-level training}} \\
        DA & 0.525 & 0.557 & 0.217 & 0.433\\
        DA (3 LPs) & 0.560 & 0.561 & 0.222 & 0.448\\
        DA + MQM & 0.540 & 0.568 & 0.262 & 0.457\\
        DA (3 LPs) + MQM & 0.553 & \bf 0.569 & 0.268 & 0.463 \\
        DA (3 LPs) concat. MQM & \bf 0.578 & 0.547 & \bf 0.278 & \bf 0.468\\\midrule
        \multicolumn{5}{c}{\textit{Final Ensemble}} \\
        Ensemble 11x & 0.568 & 0.556 & 0.223 & 0.449\\
        \bottomrule
    \end{tabular}
    \caption{Results for sentence-level QE in terms of Spearman correlation for MQM. 
    % \marcos{To save space we can try to squeeze this table within a column} \marcos{Alternatively, would be possible to join this table with the table 1? we can fit all into a single table by reporting less LPs or by using some latex tricks.}
    }
    \label{tab:results_task_1_MQM}
\end{table}

\begin{table*}[t]
    \centering
    \small    
    \begin{tabular}{l cccccc c@{} cccc}
        \toprule
        & \multicolumn{6}{c}{\bf Post-edit} & & \multicolumn{4}{c}{\bf MQM} \\
        \cmidrule{2-7}
        \cmidrule{9-12}
        \bf Method & \bf en-cs & \bf en-ja & \bf en-mr & \bf km-en & \bf ps-en & \bf avg.  & & \bf en-de & \bf en-ru & \bf zh-en & \bf avg. \\
        \midrule
        Baseline~\citep{zerva-etal-2021-ist} & 0.272 &	0.154 & 0.326 & 0.427 & 0.348 & 0.305  & & 0.176 &	0.177  & 0.065 & 0.139 \\
        \midrule
        \textit{InfoXLM as encoder} \\
        Word-level & 0.351 & 0.183 & 0.337 & 0.443 & 0.372 & 0.337 & & - & - & - & - \\
        \,\, + Sentence-level & 0.410 & 0.230 & 0.368 & 0.436 &	0.369 & 0.363 & & 0.294 &	0.256 &	0.399 &	0.316 \\
        \,\, + LP prefix  &  0.371 & 0.202 & 0.391 & 0.512 & 0.411 & 0.377 & & 0.259 &	0.440 &	0.211 & 0.303 \\
        \,\, + APEQuest \& QT21 & 0.414 & 0.245 & 0.372 & 0.494 & 0.389 & 0.383 & & 0.246 & 0.382  & 0.209 & 0.279\\
        \,\, + tuned class-weights & 0.389 & 0.218 & 0.421 & 0.499 & 0.391 & 0.384 & & 0.285 &	0.404 &	0.172 & 0.287  \\
        DA (3LPs) + MQM     & - & - & - & - & - & - & &	0.265 & 0.367 & 0.360 & 0.331 \\
        \textit{RemBERT as encoder}\\
        Word + sentence-level &	0.353 & 0.163 & 0.303 & 0.443 & 0.369 & 0.326 & & 0.262 &	0.309 &	0.147 &	0.240\\
        \,\, + LP prefix & 0.384 &	0.257 &	0.375 &	0.460 &	0.370 &	0.369 & & 0.288 & 0.356	 & 0.297 &	0.313\\
        %+ LP prefix + pretrained on APEQuest \& QT21  & \\
        \midrule
        Ensemble ``best-only'' & 0.414 & 0.245 & 0.421 & 0.512 & 0.411 & 0.401 & & 0.300 & 0.382 & 0.360 & 0.347\\
        Ensemble logits & \bf 0.438	& \bf 0.257	& \bf 0.445	& \bf 0.547	& \bf 0.430 & \bf 0.423 & & \bf 0.325	& 0.443	& 0.296 & 0.355\\
        Ensemble tags & 0.432 & 0.253 & 0.429 & 0.537 & 0.423	& 0.415 & & 0.313 & \bf 0.446 & \bf 0.408 & \bf 0.389 \\
        \bottomrule
    \end{tabular}
    \caption{Results for word-level QE in terms of MCC for the post-edit and MQM LPs. Note that in each row, we use models trained separately on the MQM and non-MQM LPs. %\nuno{I prefer the formatting of Table 5 for the rows w/ "+ ..."} \nuno{Do we have RemBERT w/o sentence-level signal?}
    }
    \label{tab:results_task_1_word_level_da}
\end{table*}

\paragraph{Word-level.} 
For the word-level task we tuned models separately for the LPs that consisted of post-edit-derived word tags and the ones consisting of MQM-derived word tags; we report the Matthew's correlation coefficient (MCC) in Table \ref{tab:results_task_1_word_level_da}. We experimented with multi-tasking by adding sentence-level supervision to the word-level task and found that it boosts performance especially for the out-of-English translations. For the non-MQM LPs we used the HTER scores as sentence level targets as we found they lead to significantly higher correlations. We can also see that using the sentence-mix and the language prefix boosted the performance for all LPs, both in the MQM and post-edit originated LPs. 
Overall, the results show further improvements when we use the HTER scores of APEQuest and QT21 as additional pretraining data, but only for specific LPs. These findings merit further investigation, since the directionality of the LPs seems to have impacted our experiments.
Finally, ensembling led to better results across all languages. Ensembling the logits led to better results for the post-edit originated LPs, while word-level ensembling helped more the MQM-originated LPs. Yet, in the submitted versions we found that the difference in performance between the three ensembling methods yielded similar results, with only 1-2\% difference, while in the averaged multilingual versions these differences were even smaller, varying less than 0.1\%.

\subsection{Explainable QE}

Since the explanations are given as continuous scores, they are evaluated against the ground-truth word-level labels in terms of the Area Under the Curve (AUC), Average Precision (AP), and Recall at Top-K (R@K) metrics only on the subset of translations that contain errors. Although R@K was considered the main metric for this task, we optimized internally for the average of all three metrics. The results are shown in Table~\ref{tab:results_task_2}.

\paragraph{Discussion.} The results highlight several contrasts between explanations for DA and MQM data: (i) while RemBERT is useful as an encoder for DA data (outperforms InfoXLM in 3 out of 5 LPs), it is outperformed by InfoXLM for all MQM LPs; (ii) the Head Mix component improves performance for DA, but it does not impact significantly the scores for MQM; and (iii) the Sparse Head Mix generally outperforms the Soft Head Mix for DA, but the trend flips for MQM. On what comes to the explainability methods, the baseline method (Attn $\times$ Norm -- scaling the attention weights by the $\ell_2$-norm of value vectors), which obtained the best results in last year's Explainable QE shared task, is outperformed by our new method ($\text{Attn} \times \text{GradNorm}$) for both DA and MQM data. Moreover, ensembling explanations from different heads brings further consistent improvements across the board for all LPs. For the zero-shot setting (\textit{en-yo}), we build an ensemble of explanations by using the heads that were more common among the ensembles for all other LPs. 
% As a preliminary strategy, we investigated the correlation between Sparse Head Mix coefficients and the AUC score of each head, and we found that 
This approach might be worth researching further, since it is possible to study the Head Mix coefficients to select good-performing attention heads. 

\begin{table*}[t]
    \centering
    \small    
    \begin{tabular}{l cccccc c@{ } cccc}
        \toprule
        & \multicolumn{6}{c}{\bf Direct Assessment} & & \multicolumn{4}{c}{\bf MQM} \\
        \cmidrule{2-7} \cmidrule{9-12}
        \bf Method & \bf en-cs & \bf en-ja & \bf en-mr & \bf km-en & \bf ps-en & \bf avg. & & \bf en-de & \bf en-ru & \bf zh-en & \bf avg. \\
        \midrule
        Baseline~\citep{treviso-etal-2021-ist}$^\dagger$ & 0.602 & 0.510 & 0.428 & 0.636 & 0.633 & 0.562 & & 0.529 & 0.552 & 0.450 & 0.510 \\
        \midrule
        % \textit{Other competitors} \\
        % HW-TSC    & 0.000 & 0.000 & 0.000 & 0.000 & 0.000 & 0.000 & 0.000 & 0.000 & 0.000 & 0.000 \\
        % f.azadi    & 0.000 & 0.000 & 0.000 & 0.000 & 0.000 & 0.000 & 0.000 & 0.000 & 0.000 & 0.000 \\
        % \midrule
        \textit{InfoXLM as encoder} \\
        % Attn $\times$ Norm (baseline)   & 0.000 & 0.000 & 0.000 & 0.000 & 0.000 & 0.000 & & 0.000 & 0.000 & 0.000 & 0.000 \\
        Attn $\times$ GradNorm          & 0.602 & 0.495 & 0.417 & 0.653 & 0.648 & 0.563 & & 0.539 & 0.559 & 0.474 & 0.524 \\
        \,\, + Soft Head Mix            & 0.600 & 0.495 & 0.426 & 0.656 & 0.653 & 0.566 & & 0.532 & 0.563 & 0.467 & 0.521 \\
        \,\, + Sparse Head Mix          & 0.604 & 0.503 & 0.421 & 0.658 & 0.660 & 0.569 & & 0.541 & 0.551 & 0.454 & 0.515 \\
        Ensemble                        & 0.641 & 0.521 & 0.440 & 0.669 & 0.667 & 0.588 & & \textbf{0.580} & \textbf{0.603} & \textbf{0.505} & \textbf{0.563} \\
        \,\, + Soft Head Mix            & 0.621 & 0.501 & 0.432 & 0.681 & 0.661 & 0.579 & & 0.567 & 0.588 & 0.504 & 0.553 \\
        \,\, + Sparse Head Mix          & \textbf{0.645} & 0.519 & \textbf{0.450} & 0.688 & 0.675 & 0.595 & & 0.574 & 0.582 & 0.484 & 0.547 \\
        \midrule
        \textit{RemBERT as encoder} \\
        Attn $\times$ GradNorm          & 0.596 & 0.511 & 0.427 & 0.675 & 0.676 & 0.577 & & 0.474 & 0.532 & 0.448 & 0.485 \\
        \,\, + Soft Head Mix            & 0.588 & 0.538 & 0.430 & 0.658 & 0.654 & 0.574 & & 0.473 & 0.529 & 0.455 & 0.486 \\
        \,\, + Sparse Head Mix          & 0.588 & 0.534 & 0.428 & 0.658 & 0.652 & 0.572 & & 0.470 & 0.530 & 0.443 & 0.481 \\
        Ensemble                        & 0.609 & 0.551 & 0.443 & \textbf{0.702} & 0.685 & 0.598 & & 0.516 & 0.554 & 0.506 & 0.525 \\
        \,\, + Soft Head Mix            & 0.613 & \textbf{0.561} & 0.448 & 0.699 & 0.692 & 0.603 & & 0.521 & 0.558 & 0.498 & 0.526 \\
        \,\, + Sparse Head Mix          & 0.620 & 0.557 & 0.447 & \textbf{0.702} & \textbf{0.691} & \textbf{0.604} & & 0.511 & 0.551 & 0.503 & 0.522 \\
        \bottomrule
    \end{tabular}
    \caption{Explainable QE task results in terms of the average of AUC, AP and R@K. $^\dagger$We used InfoXLM to compute the results for the baseline.}
    \label{tab:results_task_2}
\end{table*}

\section{Official Results}

We present the official results of our submissions alongside the results from other competitors in Section~\ref{sec:official_results_apendix} for all three tasks. For sentence-level, our submissions achieved the best results for 6/9 LPs. For word-level, we obtained the best results for 5/9 LPs. For the explainable QE track, we obtained the best results for all but two LPs (\textit{km-en} and \text{ps-en}). 
Although the critical error detection task had no other competitor for the \textit{constrained setting}, our submission vastly surpassed the organizers' baseline. 
We also obtained the best results for the multilingual settings (including and excluding \textit{en-yo}) for all tasks.
Finally, when averaging the results for all LPs, our submissions place on top for all tasks. %\marcos{this section can be removed (or moved to the appendix) to favor space}

\section{Conclusions and Future Work}

We presented the joint contribution of IST and Unbabel to the WMT 2022 QE shared task. 
We found that incorporating references during pretraining improves performance across several LPs on downstream tasks, and that jointly training with sentence and word-level objectives yields a further boost.
For Task 1, our final submissions were ensembles of models finetuned with different pretrained language models as encoders, boosting the results when compared to the previous year submission.
For Task 2, we take inspiration on the literature of explainability and propose to use gradient information in tandem with attention weights, and to further refine the impact of attention heads towards the prediction via the Head Mix component. Besides leading to better explainability performance for some LPs, this strategy is potentially useful to identify good attention heads at inference time for zero-shot LPs, and deserves more investigation. 
Overall, our submissions achieved the best results for all tasks (including Task 3) for almost all LPs by a considerable margin. 

One of the challenges of leveraging big ensembles is the burdensome weight of parameters and inference time. For future work we will extend our recent work, \textsc{Cometinho}~\cite{rei-etal-2022-searching} and explore how to effectively distill large ensembles into small and more practical QE systems.

% We have shown that the multi-head mechanism---the bedrock on which transformers are built---is able to learn the importance of tokens associated with \textsc{bad} tags. Furthermore, composing explanations in the form of attention probabilities scaled by the norm of value vectors leads to further improvements \citep{kobayashi-etal-2020-attention}. Ensembling these explanations yields the best results overall for all tested metrics on all LPs, including zero-shot ones.

\section*{Acknowledgements}
This work was supported by the P2020 program MAIA (contract 045909), by the European Research Council (ERC StG DeepSPIN 758969), and by the Funda\c{c}\~{a}o para a Ci\^{e}ncia e Tecnologia through contract UIDB/50008/2020.

\bibliography{anthology, custom}
\bibliographystyle{acl_natbib}

\appendix

\section{Data Information}
\label{sec:data}

The data used for finetuning our QE systems is shown in Table~\ref{tab:qe_data}. For DA data, we split the original development set to generate a new dev/test split, therefore the reported numbers in the table correspond to this ``internal'' dev split. %\marcos{todo: add test set info} \marcos{can someone confirm if these dev stats are actually from our internal dev split or what?}.

\begin{table}[!htb]
    \centering
    \small
    \begin{tabular}{lrrrr}
        \toprule
        & & Source & Target & Target \\
        LP & Samples & Tokens & Tokens & \textsc{ok} / \textsc{bad}   \\
        \midrule
        % train DA
        \sc train \\
        en-de & 9000 & 147870 & 153656 & 0.84 / 0.16 \\
        en-mr & 26000 & 690516 & 561371 & 0.90 / 0.10 \\
        en-zh & 9000 & 148657 & 163308 & 0.65 / 0.35 \\
        et-en & 9000 & 126877 & 185491 & 0.75 / 0.25 \\
        ne-en & 9000 & 135205 & 181707 & 0.41 / 0.59 \\
        ro-en & 9000 & 154538 & 167471 & 0.71 / 0.29 \\
        ru-en & 9000 & 104423 & 132006 & 0.85 / 0.15 \\
        si-en & 9000 & 141283 & 166914 & 0.42 / 0.58 \\
        % train MQM
        en-de$^\dagger$ & 54681 & 1571090 & 1926444 & 0.90 / 0.10 \\
        en-ru$^\dagger$ & 15628 & 312185 & 354871 & 0.95 / 0.05 \\
        zh-en$^\dagger$ & 75327 & 134165 & 2789907 & 0.87 / 0.13 \\
        \midrule
        % dev DA
        \sc dev \\
        en-de & 500 & 8262 & 8555 & 0.84 / 0.16 \\
        en-mr & 500 & 13803 & 11216 & 0.91 / 0.09 \\
        en-zh & 500 & 8422 & 9302 & 0.75 / 0.25 \\
        et-en & 500 & 7081 & 10257 & 0.73 / 0.27 \\
        ne-en & 500 & 7542 & 10247 & 0.38 / 0.62 \\
        ro-en & 500 & 8550 & 9202 & 0.78 / 0.22 \\
        ru-en & 500 & 5984 & 7511 & 0.84 / 0.16 \\
        si-en & 500 & 7866 & 9415 & 0.41 / 0.59 \\
        en-cs & 500 & 10302 & 9302 & 0.75 / 0.25 \\
        en-ja & 500 & 10354 & 13287 & 0.73 / 0.27 \\
        km-en & 495 & 9015 & 8843 & 0.45 / 0.55 \\
        ps-en & 500 & 13463 & 12160 & 0.51 / 0.49 \\
        % dev MQM
        en-de$^\dagger$ & 503 & 10535 & 12454 & 0.96 / 0.04 \\
        en-ru$^\dagger$ & 503 & 10767 & 11911 & 0.91 / 0.09 \\
        zh-en$^\dagger$ & 509 & 980 & 19192 & 0.98 / 0.02 \\
        \bottomrule
    \end{tabular}
    \caption{DA and MQM ($\dagger$) data for all LPs.}
    \label{tab:qe_data}
\end{table}

\section{Official Results}\label{sec:official_results_apendix}

\paragraph{Critical Error Detection.} Submissions for this task were evaluated in terms of ranking using R@K and MCC as metrics. In Table~\ref{tab:results_task_3}, we report only MCC scores as it was the main metric for this task.

\begin{table}[!htb]
    \centering
    \small
    \begin{tabular}{lcc}
        \toprule
        \bf Method & \bf en-de & \bf pt-en  \\
        \midrule
        Baseline                   & 0.0738 & -0.0013 \\
        InfoXLM finetuned on DAs   & 0.5641 & 0.7209 \\
        \bottomrule
    \end{tabular}
    \caption{Official results for the Critical Error Detection task in terms of MCC.}
    \label{tab:results_task_3}
\end{table}

\paragraph{QE and Explainable QE.} Table~\ref{tab:results_task_1_and_2} shows the official results for sentence-level QE (top) in terms of Spearman's correlation, word-level QE (middle) in terms of MCC, and explainable QE (bottom) in terms of R@K. 

\begin{table*}
    \centering
    \small
    \begin{tabular}{l cccccccc c@{ } ccc}
        \toprule
        & \multicolumn{8}{c}{\bf Direct Assessment} & & \multicolumn{3}{c}{\bf MQM} \\
        \cmidrule{2-9} \cmidrule{11-13}
        \bf Team & \bf en-cs & \bf en-ja & \bf en-mr & \bf en-yo & \bf km-en & \bf ps-en & \bf all & \bf all/yo & & \bf en-ru & \bf en-de & \bf zh-en\\
        \midrule
        \textit{Sentence-level QE} \\
        Baseline                    & 0.560 & 0.272 & 0.436 & 0.002 & 0.579 & 0.641 & 0.415 & 0.497 & & 0.333 & 0.455 & 0.164 \\
        Alibaba                     & -     & -     & -     & -     & -     & -     & -     & -     & & 0.505 & 0.550 & 0.347 \\
        NJUQE                       & -     & -     & 0.585 & -     & -     & -     & -     & -     & & 0.474 & \bf 0.635 & 0.296 \\
        Welocalize                  & 0.563 & 0.276 & 0.444 & -     & 0.623 & -     & 0.448 & 0.506 & & -     & -     & -     \\
        joanne.wjy                  & 0.635 & 0.348 & 0.597 & -     & 0.657 & 0.697 & -     & 0.587 & & -     & -     & -     \\
        HW-TSC                      & 0.626 & 0.341 & 0.567 & -     & 0.509 & 0.661 & -     & -     & & 0.433 & 0.494 & \bf 0.369 \\
        Papago                      & 0.636 & 0.327 & \bf 0.604 & 0.121 & 0.653 & 0.671 & 0.502 & 0.571 & & 0.496 & 0.582 & 0.325 \\
        IST-Unbabel                 & \bf 0.655 & \bf 0.385 & 0.592 & \bf 0.409 & \bf 0.669 & \bf 0.722 & \bf 0.572 & \bf 0.605 & & \bf 0.519 & 0.561 & 0.348 \\
        \midrule
        \textit{Word-level QE} \\
        Baseline                    & 0.325 & 0.175 & 0.306 & 0.000 & 0.402 & 0.359 & 0.235 & 0.257 & & 0.203 & 0.182 & 0.104 \\
        NJUQE                       & -     & -     & 0.412 & -     & 0.421 & -     & -     & -     & & 0.390 & \bf 0.352 & 0.308 \\
        HW-TSC                      & 0.424 & \bf 0.258 & 0.351 & -     & 0.353 & 0.358 & -     & 0.218 & & 0.343 & 0.274 & 0.246 \\
        Papago                      & 0.396 & 0.257 & \bf 0.418 & 0.028 & \bf 0.429 & 0.374 & 0.317 & 0.343 & & 0.421 & 0.319 & 0.351 \\
        IST-Unbabel                 & \bf 0.436 & 0.238 & 0.392 & \bf 0.131 & 0.425 & \bf 0.424 & \bf 0.341 & \bf 0.361 & & \bf 0.427 & 0.303 & \bf 0.360 \\
        \midrule
        \textit{Explainable QE} \\
        Baseline                    & 0.417 & 0.367 & 0.194 & 0.111 & 0.580 & 0.615 & 0.381 & 0.435 & & 0.148 & 0.074 & 0.048 \\
        f.azadi                     & -     & -     & -     & -     & 0.622 & 0.668 & -     & -     & & -     & -     & -     \\
        HW-TSC                      & 0.536 & 0.462 & 0.280 & -     & \bf 0.686 & \bf 0.715 & -     & 0.535 & & 0.313 & 0.252 & 0.220 \\
        IST-Unbabel                 & \bf 0.561 & \bf 0.466 & \bf 0.317 & \bf 0.234 & 0.665 & 0.672 & \bf 0.486 & \bf 0.536 & & \bf 0.390 & \bf 0.365 & \bf 0.379 \\
        
        \bottomrule
    \end{tabular}
    \caption{Official results for sentence-level QE (top) in terms of Spearman's correlation, word-level QE (middle) in terms of MCC, and explainable QE (bottom) in terms of R@K. We estimated the numbers of \textit{en-yo} for teams that did not submit to \textit{en-yo} directly but still submitted to all other LPs and to the \textit{multilingual} (all) category.}
    \label{tab:results_task_1_and_2}
\end{table*}

\end{document}